\documentclass{article} 
\usepackage[final]{colm2026_conference}

\usepackage{microtype}
\usepackage{hyperref}
\usepackage{url}
\usepackage{graphicx}
\usepackage{subcaption}
\usepackage{adjustbox}
\usepackage{booktabs}  
\usepackage{multirow}  
\usepackage{colortbl}  
\definecolor{colOutput}{RGB}{174,214,241}   
\definecolor{colPhase}{RGB}{130,204,174}    
\definecolor{colPlan}{RGB}{248,196,113}     
\definecolor{colAsync}{RGB}{195,155,211}    
\usepackage{pifont}

\usepackage{lineno}

\definecolor{darkblue}{rgb}{0, 0, 0.5}
\hypersetup{colorlinks=true, citecolor=darkblue, linkcolor=darkblue, urlcolor=darkblue}

\title{Position: AI Agents in Scientific Teams Should Be Studied as Human-Agent Systems}

\author{Patrick Emami\thanks{patrick.emami@nlr.gov}, Jared D. Willard, Saumya Sinha, Truc Nguyen, Andrew Glaws \\
National Laboratory of the Rockies \\
\And
Sameera Horawalavithana\thanks{yasanka.horawalavithana@pnnl.gov}, Gihan Panapitiya, Bruno Jacob, Siddhisanket Raskar \\
Pacific Northwest National Laboratory \\
\And
Nithin Somasekharan, Ling Yue, Shaowu Pan \\
Rensselaer Polytechnic Institute \\
\And 
Brian Lu, Jason Eisner \\
Johns Hopkins University
}

\begin{document}

\ifcolmsubmission
\linenumbers
\fi

\maketitle

\begin{abstract}
Large language model-based agents are increasingly deployed as collaborators in scientific discovery yet most current work focuses on the autonomous capabilities of ``AI Scientists''.
We argue that this overlooks the social aspects of scientific teamwork, and that studying AI Scientists as human-agent systems (HAS)—where the unit of analysis is the human-agent pair—is both underexplored and undervalued. We establish these points through literature and empirical analysis, and highlight recent incidences and studies which show that deploying agents in science without accounting for human-agent dynamics introduces near-term risks, including reduced diversity of scientific inquiry. Through analysis of real-world case studies, we show that scientists and agents can augment each other's capabilities. We call for new research that adopts the HAS lens to develop mathematical frameworks for understanding and fostering human-AI synergy in scientific discovery.
 \end{abstract}

\section{Introduction}\label{sec:intro}
Modern science is a social endeavor that increasingly involves working in teams~\citep{fortunato2018science,NAP29043}.
Collaboration helps scientists achieve their lofty goals by bringing experts with diverse scientific backgrounds and experiences together, combining their strengths to collectively answer highly complex questions.
Science teams vary widely in their form and function, from small university laboratories to large institutions that coordinate hundreds of members.
Meta-analysis of scientific teamwork has found that working in a team increases impact; in fact, teams are 6.3 times more likely to produce a highly cited paper than a solo author~\citep{wuchty2007increasing}.

Today, scientific teamwork is undergoing a major shift as large language model (LLM)-based agents---systems that perceive, reason, and act in an environment, which we refer to simply as ``agents''---are being deployed not only as tools for scientists, but as collaborators in scientific processes. 
The introduction of agents into scientific teamwork sits within broader trends of 1) increasing prevalence and advantage of scientific teamwork~\citep{hall2018science} and 2) increasing levels of automation in science.
Position papers envision AI Scientists as evolving towards higher levels of autonomy with less and less reliance on human judgment and oversight~\citep{gao2024empowering,liu2025foundation,zhang2025scaling,zhang2025evolving,wei2025ai,xie2025far,zhou2025autonomous,zheng2025automation,ren2025towards,tie2026autoresearch}.
In Section~\ref{sec:related-work}, we analyze current agentic systems adapted for scientific discovery, which we call \textbf{AI Scientists}, finding that the majority of recent work only considers a narrow form of human participation: setting goals and verifying work. Fewer systems support interaction geared at sustained, iterative, and joint collaboration typical of human-human teamwork. 
This literature has largely been isolated from the socio-technical realities of science.

If AI Scientists are to be effectively integrated into scientific teams, it is timely that we also begin to study them as human-agent systems (HAS).
An HAS considers as core components: agents, human participants, an interface that facilitates human-agent interaction through actions, and the environment which the HAS is situated in~\citep{haupt2025position,zou2025llm,luo2026centaureval}. 
In HAS, the unit of study is the pair: \emph{humans and agents}.
\textbf{Our position is that studying AI Scientists as HAS is underexplored and undervalued, and that new research which adopts the lens of HAS is needed to develop an understanding of \emph{human-AI synergy} in scientific discovery settings.}

We argue that adopting an HAS lens can help us better understand---and thus potentially mitigate---near-term risks of deploying AI Scientists at scale.
Evidence of the risks of AI in science is accruing; for example, a recent study showed that AI-augmented research produces three times as many papers at a 5\% reduction in the diversity of explored topics~\citep{hao2026artificial}.
In Section~\ref{sec:risks}, we provide further evidence from recent incidents and studies, and discuss the importance of framing AI Scientists as HAS to mitigation efforts.

We also argue that an HAS perspective presents a path to exploring new scientific questions and discovering new solutions inaccessible to humans alone and AI alone.
In Section~\ref{sec:case-studies}, we analyze recent real-world case studies of scientists co-working alongside agents that qualitatively illustrate how humans and agents have augmented each other's outputs, achieving a level of human-AI synergy that tangibly improved scientific outcomes.
Finally, we raise new questions about the study of agents in scientific discovery motivated by the HAS lens in Section~\ref{sec:questions}, and discuss alternative views in Section~\ref{sec:alternatives}.

\section{AI Scientists as HAS: A Review and Preliminary Experiment}\label{sec:related-work}

\begin{table*}[!t]
\footnotesize
\centering
\caption{A representative set of agents categorized by the primary channel for human feedback (row-color). The three feedback dimensions (Type, Granularity, and Phase) follow~\citet{zou2025llm}'s taxonomy of human feedback in LLM HAS. Systems can support multiple channels to varying degrees (see Table~\ref{tab:when}). See Section~\ref{sec:related-work} for definitions.}
\label{tab:relwork}
\setlength{\tabcolsep}{5pt}
\begin{tabular}{lp{1.8cm}p{1.7cm}p{1.6cm}}
\toprule
\textbf{System} & \textbf{Type} & \textbf{Granularity} & \textbf{Phase} \\
\midrule
\rowcolor{colOutput!70} \multicolumn{4}{l}{\textbf{Artifact-level}} \\
\rowcolor{colOutput!30} AI Scientist v2~\citep{yamada2025ai}            & Evaluative & Coarse & Post-task \\
\rowcolor{colOutput!30} Kosmos~\citep{mitchener2025kosmos}              & Evaluative & Coarse & Post-task \\
\rowcolor{colOutput!30} 
\rowcolor{colOutput!30} CycleResearcher~\citep{weng2024cycleresearcher} & Evaluative & Coarse & Post-task \\
\rowcolor{colOutput!30} VirSci~\citep{su2025many}                       & Evaluative & Coarse & Post-task \\
\rowcolor{colOutput!30} AgentRxiv~\citep{schmidgall2025agentrxiv}       & Evaluative & Coarse & Post-task \\
\rowcolor{colOutput!30} AI-Researcher~\citep{tang2026ai}                & Evaluative & Coarse & Post-task \\
\rowcolor{colOutput!30} DeepScientist~\citep{weng2025deepscientist}     & Evaluative & Coarse & Post-task \\
\rowcolor{colPhase!70}  \multicolumn{4}{l}{\textbf{Discovery-phase-level}} \\
\rowcolor{colPhase!30}  InternAgent~\citep{team2025internagent}         & Evaluative & Coarse & Mid-task \\
\rowcolor{colPhase!30}  Co-Scientist~\citep{gottweis2025towards}        & Evaluative & Coarse & Mid-task \\
\rowcolor{colPhase!30}  AgentLaboratory~\citep{schmidgall2025agent}     & Evaluative & Coarse & Mid-task \\
\rowcolor{colPhase!30}  
\rowcolor{colPhase!30} Denario~\citep{villaescusa2025denario}          &  Corrective & Fine & Mid-task \\
\rowcolor{colPlan!70}   \multicolumn{4}{l}{\textbf{Research-plan-level}} \\
\rowcolor{colPlan!30}   MAPPS~\citep{zhou2025toward}                    & Guidance   & Fine   & Mid-task \\
\rowcolor{colPlan!30}   El Agente Q~\citep{zou2025agente}               & Guidance   & Fine   & Pre-task \\
\rowcolor{colPlan!30}   Organa~\citep{darvish2025organa}                & Guidance   & Fine   & Pre-task \\
\rowcolor{colPlan!30}   SciSciGPT~\citep{shao2025sciscigpt}            & Guidance   & Fine   & Pre-task \\
\rowcolor{colAsync!70}  \multicolumn{4}{l}{\textbf{Asynchronous steering \& interruption}} \\
\rowcolor{colAsync!30}  FreePhDLabor~\citep{li2025build}               & Corrective & Fine   & Mid-task \\
\rowcolor{colAsync!30} EvoScientist~\citep{lyu2026evoscientist}        & Corrective & Fine & Mid-task \\
\rowcolor{colAsync!30}
OmniScientist~\citep{shao2025omniscientist}    & Evaluative & Coarse & Mid-task \\
\rowcolor{colAsync!30}
ScienceClaw~\citep{wang2026autonomous} & Guidance & Coarse & Mid-task \\
\bottomrule
\end{tabular}
\end{table*}
\begin{table}[]
\caption{A preliminary evaluation of feedback-seeking behavior in a ReAct GPT-5-mini agent on the End-to-End Discovery (E2E) AstaBench benchmark. We compare three conditions: (1) the AstaBench planning-centric ReAct agent harness, (2) adding an \texttt{ask\_user} tool and the HAS human feedback taxonomy to the system prompt as \emph{feedback encouragement}, and (3) adding explicit guidance to always call the \texttt{ask\_user} tool before every action. Condition (3) acts as an upper bound on agent disengagements. We use LLM-as-a-judge to classify  questions under the taxonomy and show dimension-wise percentages (counts across 10 E2E development set tasks normalized by total questions asked), averaged across three judges for robustness. \textbf{Key findings:} The agent without explicit per-step instruction \emph{does not know when to ask}, e.g., it prefers to make assumptions about how to handle ambiguous situations when critical resources are unavailable. With per-step enforcement, the dominant feedback type is evaluative, generally occurring mid-task. Second most common behavior is guidance-based, e.g., asking the user their preference given options, occurring heavily in the pre-task phase. Corrective feedback is rarely requested.}
\label{tab:when}
\centering
\footnotesize
\begin{adjustbox}{max width=\columnwidth}
\begin{tabular}{@{}ccccccccccccc@{}}
\toprule
\multicolumn{3}{c}{Condition} & \texttt{ask\_user} calls & \multicolumn{3}{c}{Type} & \multicolumn{3}{c}{Phase} & \multicolumn{2}{l}{Granularity} & \multicolumn{1}{c}{} \\ \midrule
\multicolumn{1}{l}{\texttt{ask\_user} tool} & \multicolumn{1}{l}{Feedback taxonomy} & \multicolumn{1}{l}{Per-step} & & E & G & \multicolumn{1}{c|}{C} & I & D & \multicolumn{1}{c|}{P} & C & \multicolumn{1}{c|}{F} & E2E Score \\ \midrule
 &  &  & 0 & 0 & 0 & 0 & 0 & 0 & 0 & 0 & 0 & 0.39 \\
\ding{51} & \ding{51} &  & 1 &  0 & 1.0 & 0 & 1.0 & 0 & 0 & 0 & 1.0 & 0.36 \\
\ding{51} &\ding{51} &\ding{51} & 92 & 0.71 & 0.25 & 0.04 & 0.35 & 0.45 & 0.20 & 0.68 & 0.32 & 0.24 \\ \bottomrule
\end{tabular}
\end{adjustbox}
\end{table}
We adopt an HAS lens to review the literature on AI Scientists, with the goal of broadly understanding how current systems interact with human scientists.
Our review categorizes AI Scientist systems based on the primary channel which human feedback enters the scientific discovery loop: artifact-level feedback (Section~\ref{sec:rel:output-level}), discovery phase-level feedback (Section~\ref{sec:rel:phase}), research planning feedback (Section~\ref{sec:rel:plan}), and asynchronous steering \& interruption (Section~\ref{sec:relwork:mem}).
We adopt three human feedback dimensions from~\citet{zou2025llm}'s taxonomy for LLM HAS in Table~\ref{tab:relwork} to further organize systems: \textbf{Type}---\emph{Evaluative} (E): assessment of output quality, \emph{Guidance} (G): instructions, demonstrations, or critiques, or \emph{Corrective} (C): user edits or fixes, \textbf{Granularity}---\emph{Coarse} (C): Single assessment for an entire output or outcome, \emph{Fine} (F): Targeted step-wise feedback, and \textbf{Phase}---\emph{Initial setup \& goal} (I): Pre-task, \emph{During task execution} (D): Mid-task, \emph{Post-task eval \& refinement} (P).

\subsection{Artifact-level feedback}\label{sec:rel:output-level}

Early AI Scientist systems automate the full discovery pipeline—ideation, experimentation, manuscript preparation—requiring humans only to specify a research question and evaluate final outputs~\citep{lu2024ai,yamada2025ai,ghafarollahi2025sparks,tang2026ai,weng2025deepscientist}.
Tree-search and Bayesian-optimization extensions improve exploration and reduce hallucinations by filtering hypotheses with computational validation~\citep{yamada2025ai,weng2025deepscientist,liu_aigs_2024,yu2025alpharesearch}.
Tighter internal feedback loops feed experiment results back into ideation~\citep{yuan2025dolphin} or pair the agent with a peer-review trained reviewer model for iterative refinement~\citep{weng2024cycleresearcher}, reducing reliance on human review before final evaluation.
Multi-agent variants organize LLM agents into collaborative teams~\citep{su2025many}, structured multi-stage workflows~\citep{tang2026ai}, and swarm reasoning over knowledge graphs~\citep{ghafarollahi2024sciagents}; agent preprint servers enable sharing results for cross-lab iteration~\citep{schmidgall2025agentrxiv}.
Autoresearch demonstrates that a simple coding harness and LLM can serve a similar role when also equipped with domain-specific tools and an iterative evaluation loop~\citep{karpathy2025aautoresearch}.
In all cases, humans remain as initiator and final evaluator.

\subsection{Discovery-phase-level feedback}\label{sec:rel:phase}

A large class of AI Scientists permit humans to intervene after one or more phases of a sequential discovery workflow, canonically: literature review, ideation and hypothesis generation, experimentation, then manuscript preparation and review.
Across these systems, human feedback is primarily evaluative and corrective; granularity level varies from coarse (fixed review gates) to fine (editable phase-level memory).

Multi-agent systems assign phase-specific roles to agents and expose phase-level review gates where scientists provide high-level feedback~\citep{schmidgall2025agent,gottweis2025towards,ghareeb2025robin,villaescusa2025denario}.
Denario's Markdown-based memory system exposes each phase-specific module's inputs and outputs, enabling fine-grained oversight by editing intermediate state before resuming execution~\citep{villaescusa2025denario}.
Transparency and rewind mechanisms (e.g., linking generated figures to source code, reverting to earlier phases) further increase steerability~\citep{ifargan2025autonomous,wang2026autonomous,liu2026last}.

Phase-specific feedback targets common failure modes such as hypothesis vagueness.
Strategies range from high-level textual feedback on ideas~\citep{team2025internagent,gottweis2025towards,ni2024matpilot}, to fine-grained feedback on complex outputs like 3D crystal structures\citep{qi2024metascientist}, to idea down-selection from ranked pools~\citep{yamada2025ai,jansen2025codescientist}.
Google's Co-Scientist for hypothesis generation adds a generate-debate-evolve tournament over hypotheses, with scientists supplying objectives upfront and steering between rounds—yielding experimentally validated discoveries in drug repurposing and antimicrobial resistance~\citep{gottweis2025towards}. 

\subsection{Research-plan-level feedback}\label{sec:rel:plan}

Many scientific activities do not fit the sequential workflows of artifact-level and phase-level systems, which offer limited visibility into low-level agent actions and restrict human involvement to fixed feedback gates.
Planning-centric AI Scientists address this via a plan-critique-execute paradigm, generating high-level plans and incorporating human feedback before execution~\citep{he2026steer}.
Domain-specific systems constrain the planning action space to a narrow set of tools and workflows~\citep{zou2025agente,huang2025biomni,darvish2025organa,shao2025sciscigpt,zhou2025toward}, with feedback granularity ranging from pre-execution plan critique~\citep{darvish2025organa} to per-step approval~\citep{zhou2025toward}.
The human role here is primarily \emph{guidance oriented}: scientists proactively shape the agent's course of action through collaborative planning rather than reacting to outputs.


\subsection{Feedback via asynchronous steering \& interruption}\label{sec:relwork:mem}

A minority of AI Scientists offer fine-grained, asynchronous human interaction during phased discovery workflows.
FreePhDLabor~\citep{li2025build} supports real-time interruption by writing corrective feedback into persistent agent memory.
ScienceClaw~\citep{wang2026autonomous} enables scientists to inspect and redirect agents by commenting on the Infinite platform.
EvoScientist~\citep{lyu2026evoscientist} adopts a human-on-the-loop paradigm with self-evolving memory that learns user preferences across sessions.
OmniScientist~\citep{shao2025omniscientist} aims at seamless human participation across pipeline phases through a collaborative research protocol, peer-review and the human-voted ScienceArena platform.
Coding harnesses also support mid-run interruption, but only with continuous monitoring or agents that know when to ask for help (Table~\ref{tab:when}) an open challenge even for frontier LLMs~\citep{gulati2026ask}. 
Asynchronous steering and interruption offers the highest level of human control by enabling intervention at arbitrary points rather than only at phase or output boundaries.

\subsection{Preliminary empirical support}

Our synthesis of the literature reveals that most AI Scientists assign humans supervisory roles—scoping, auditing, approval—\emph{treating discovery as an isolated optimization problem weakly tied to human-led research practices}.
Artifact-level feedback is the most numerous; discovery-phase-level systems that gate input at workflow boundaries form the next tier.
Planning-centric systems constitute a smaller but growing class, and only a handful support flexible asynchronous interaction.

To strengthen these findings, we also conduct a preliminary experiment on a representative planning-based AI Scientist, the ReAct-based agent harness and GPT-5-mini from the AstaBench AI Scientist benchmark suite~\citep{bragg2025astabench}). We compare variants provided with an \texttt{ask\_user} tool~\citep{gulati2026ask} and increasing levels of feedback guidance. Questions are responded with an automatic continuation prompt (the AstaBench default).
The results on 10 End-to-End Discovery tasks in Table~\ref{tab:when} suggest there is room for improving the ability of agents to understand when to ask for feedback and how to ask for fine-grained, corrective help in challenging, end-to-end discovery settings.
We expand later on the potential benefits of humans and AI Scientists mutually correcting each other's mistakes in Section~\ref{sec:case-studies}.
While 10 tasks may appear nominally small, each task represents a full, computationally intensive idea-to-report discovery workflow, providing a deep rather than broad measure of agent behavior.

\section{Studying AI Scientists as HAS can assist near-term risk mitigation}\label{sec:risks}

Deploying AI Scientists in scientific teams raises pressing epistemic, existential, and ethical questions.
We argue that a HAS lens helps center human participation in discovery loops, aiding near-term risk mitigation.
While future AI Scientists may transcend current limitations, our emphasis is on near-term risks exacerbated when humans offload deliberative processes to AI rather than participating as teammates.

\subsection{Can AI Scientists reliably produce accurate and unbiased scientific knowledge?}

LLMs notoriously \textbf{hallucinate}~\citep{xie2025far,zhang2025evolving,liu2025foundation}: a post-hoc review found 51 \emph{accepted} NeurIPS 2025 papers containing 100 confirmed hallucinated citations\footnote{\url{https://gptzero.me/news/neurips/}}; ACL 2026 similarly flagged over 100 accepted papers citing non-existent literature\footnote{\url{https://2026.aclweb.org/acl_statement/}}.
More broadly, LLMs and agents are prone to fabricating data or results when faced with demanding tasks~\citep{miyai2026jr,luo2025more} (also observed in our E2E experiment in Table~\ref{tab:when}).
A related concern is \textbf{bias proliferation}: LLMs may over-represent certain paradigms while under-representing novel ideas~\citep{liu2025foundation}.
GPT-4-generated college admissions essays were collectively less novel than human-written ones~\citep{moon2025homogenizing}, with worrisome implications for homogenization of scientific inquiry~\citep{messeri2024artificial,hao2026artificial}.

An HAS lens calls for systems that maintain human oversight in the near term to detect hallucinations and biases.
End-of-process review (Section~\ref{sec:rel:output-level}) is inadequate; asynchronous interaction at multiple granularities---including audits of trace logs and code---can surface issues earlier~\citep{luo2025more}, if accounting for cognitive load introduced by the speed and scale of AI.

\subsection{How will AI Scientists affect our scientific institutions and human expertise?}\label{sec:risks:existential}

AI Scientists' output volume strains \textbf{peer review}, which is already overloaded: NeurIPS and ICLR submissions increased 10.4$\times$ from 2014 to 2024 without commensurate reviewer growth~\citep{wei2025aib,xie2025far,zhang2025scaling}.
They also expand the \textbf{attack surface} of scientific software and data.
Backdoor attacks can induce LLMs to produce malicious code, and poisoning crowd-sourced knowledge bases like Wikipedia can enable such attacks~\citep{souly_poisoning_2025,carlini_poisoning_2024}.
Perhaps most consequentially, over-reliance may cause \textbf{de-skilling} in scientists, especially among junior researchers~\citep{xie2025far,zhang2025evolving}. A recent study shows dependence on AI coding assistants reduces conceptual understanding~\citep{shen2026ai}. 
Younger users show higher susceptibility to cognitive offloading~\citep{gerlich2025ai}.

The HAS lens pushes us to maintain institutional health as we deploy AI Scientists.
Automated governance---red teaming~\citep{ge-etal-2024-mart}, adversarial evaluations, monitoring, and provenance tracking---can address misuse and security.
The de-skilling question points more directly to HAS design: \emph{how we interact with AI systems matters for skill formation}~\citep{shen2026ai}, making HAS research an important lever for preserving the deliberate practice through which scientists develop competencies.

\subsection{Who is responsible when AI Scientists cause harm or enable misuse?}

When AI Scientists facilitate research in harmful or sensitive areas, \textbf{accountability} is unclear.
NIST recognizes AI as uniquely exacerbating risks of enhancing chemical, biological, radiological, or nuclear weapons capabilities~\citep{ai2024artificial}. 
Legal liability for AI harms remains rapidly evolving.
LLM-based systems also heighten concerns about \textbf{unintentional plagiarism}: even when text is not verbatim, determining originality is difficult. An expert review found 36\% of LLM and AI Scientist-generated papers contained noticeable plagiarism~\citep{gupta2025all}.
Research on whether human involvement at consequential decision points preserves accountability~\citep{rahwan2019machine} would benefit both scientific quality and safety.
Designing human-agent interfaces with natural intervention opportunities and plagiarism detection tools that do not overburden peer review are important topics for future study.

\section{The potential of human-AI synergy in science: case studies}\label{sec:case-studies}

We now turn to human-AI synergy, a general goal of any HAS.
In HAS, information and capability asymmetries are a key driver of effective collaboration~\citep{hemmer2025complementarity}, creating opportunities for humans and AI to complement each other and outperform either member alone. Prior work has formalized synergy as complementary team performance in classification decision support, combining human and AI predictions to reduce errors~\citep{bansal_wu_zhou_fok_nushi_kamar_ribeiro_weld_2021, steyvers2022bayesian}. Empirically, complementarity appears rarely~\citep{vaccaro2024combinations}, though this may reflect structural barriers: incentives favor autonomy research over HAS, and costly human studies limit rigorous evaluation~\citep{haupt2025position}. A weaker but still important form is human augmentation, where AI assists people without requiring proof that AI cannot do better alone. Both complementarity and augmentation have the potential to greatly benefit scientific discovery. 

Each case study explored next is drawn from a published account of a working scientist collaborating with an agent on a concrete scientific task.
Together they illustrate patterns of mutual augmentation: humans contributing conceptual framing, domain judgment, and epistemic verification; agents compressing implementation while also contributing at the conceptual level.






\subsection{Case study \#1: ``Vibe-coding'' a minor complexity-theory result with Gemini 3 Pro}\label{sec:case2}

This case study, obtained from the collection by~\citet{woodruff2026accelerating}, concerns a complexity theorist collaborating with Gemini 3 Pro inside Google Antigravity (a \LaTeX-integrated IDE) to write up a theorem they had not previously had time to publish. 

\textbf{How AI augmented humans}: Given an initial prompt to plan and draft the paper in the human author's writing style, Gemini generated a structured roadmap (\texttt{plan.md}) and a draft manuscript (\texttt{paper.tex}), then iteratively expanded proofs and exposition over eight prompt turns. This reduced the overhead of assembling the proof details on paper plus drafting and organizing a publishable write-up of a result that might otherwise have remained informal.

\textbf{How humans augmented AI}: Human expertise was decisive at key steering points. Early in the process, the human expert identified missing technical context (including relevant prior work and a search-vs.-decision framing) and corrected an erroneous assumption in a lemma. The interaction succeeded because an expert could diagnose omissions and provide high-quality feedback; without that expertise, the same workflow would likely fail or produce a misleading paper.

\textbf{The synergy}: The workflow resembles an expert-plus-junior-collaborator pattern, where the human sets  the research direction and quality standards while the agent accelerates drafting and execution. 
The case illustrates synergy that enabled a ``minor-result'' publication, but it also surfaces a risk raised in Section~\ref{sec:risks}: if widely adopted, this interaction mode may homogenize scientific writing and may be especially risky for early-career scientists who still need to develop core writing and reasoning skills through deliberate practice.

\subsection{Case study \#2: AI-assisted reduced physics modeling of thermonuclear burn propagation with GPT-5}\label{sec:case3}

This case study, from~\citet{bubeck2025early}, concerns an expert collaborating with GPT-5 through the ChatGPT interface to develop a reduced physics model of thermonuclear burn initiation and propagation. The exercise spanned four linked activities: (1) setting up a simplified PDE-based model for physical intuition, (2) implementing numerical simulation, (3) designing numerical experiments to expose key sensitivities, and (4) developing a theory to explain the simulation outcomes. 
The expert reported that the strongest outcome of this exercise was not time saving; rather it was (4) a theory-level explanation tying known physical relations into a closed-form predictor, used to verify that the numerically elicited trends were physically coherent.

\textbf{How AI augmented humans}: GPT-5 helped translate a human-provided concept into a concrete computational workflow. It drafted a PDE representation, incorporated domain-familiar physical ingredients, discretized the system, and scaffolded an optimization routine. 

\textbf{How humans augmented AI}: Human expertise was essential in the numerics-heavy phase, where physically meaningful parameterization required iterative debugging. The human had to manually tune hot-spot and cold-fuel conditions, conductivities, alpha-particle stopping assumptions, and profile designs to obtain physically interesting behavior. Crucially, the human also provided epistemic checks: identifying when outputs were still noise, catching oversimplifications, and rejecting silent substitutions of full numerical solves with coarse approximations.

\textbf{The synergy}: This interaction resembled expert-plus-accelerated-expert-implementer collaboration.
Work that might otherwise have required prolonged coordination across multiple human experts was compressed into a single focused session. 
However, GPT-5 exhibited familiar failure modes as Case \#1: premature confidence and smoothing over thorny issues, requiring expert oversight to address.

\section{Human-AI synergy in scientific discovery: open questions}\label{sec:questions}

Can complementary scientific team performance lead to discoveries that neither humans nor agents would have made working alone?
This section builds on the risks identified in Section~\ref{sec:risks} and case studies (Section~\ref{sec:case-studies}) to motivate new research questions.

To help guide this discussion, we introduce a toy model of the overall utility obtained by a HAS on a scientific discovery task defined within a simple decision theoretic framework for human-agent collaboration inspired by~\citet{shao2024collaborative}. 
The utility of the team with humans $H$ and agents $A$ consists of just two terms: the \textbf{collaboration advantage} $\texttt{CA}(H,A)$ and the \textbf{collaboration disadvantage} $\texttt{CD}(H,A)$.
$\texttt{CA}(H,A)$ is an advantage function that measures the team performance against the best member of the team, (i.e., \emph{complementarity}, similar to the \emph{synergy gap} in multi-agent teams~\citep{pappu2026multi}).
The term $\texttt{CD}(H,A) \geq 0$ measures the cost of collaborating, capturing additional costs paid by the team due to communication and coordination overhead (e.g., tokens spent on messages, additional cognitive load incurred by human team members).
Our toy model of overall utility 
\begin{equation} \texttt{U}(H,A) =
\texttt{CA}(H,A) - \texttt{CD}(H,A)\label{eq:utility}
\end{equation} 
suggests a goal of maximizing $\texttt{CA}(H,A)$ while minimizing $\texttt{CD}(H,A)$.

\subsection{Which factors of scientific discovery environments increase CA or reduce CD?} 

This question explores how  attributes of  scientific discovery environments relate to the overall utility $\texttt{U}(H,A)$ achieved by a fixed scientific team. 
Looking at the environments in our case studies (Section~\ref{sec:case-studies}), we start to see common factors.
\begin{itemize}
    \item Task outputs tend to be subjective research artifacts such as papers, theorems, or novel abstractions.
Here, a human expert's ability to quickly judge the work, built up over years of experience, is valuable.
    \item Tasks tend to have long-horizons and are open-ended and curiosity-driven, consisting of sub-tasks that produce research artifacts which may initially appear weakly related to the original desired outcome.
    \item Tasks may rely on limited partial observations to infer underlying structure. 
\end{itemize}
These desiderata contrast with scientific environments where \texttt{CD} is likely to overwhelm \texttt{CA}: tasks where agent trajectories and outputs can be automatically verified~\citep{ye2026evaluation} or tasks where abundant training data exists (e.g., protein structure prediction~\citep{abramson2024accurate}).
Understanding which factors impact overall utility at the environment level has implications for benchmark design as well as downstream applications.

\subsection{Which dimensions of human-agent scientific teamwork increase CA or reduce CD?}
LLMs have changed how scientists interface with AI dramatically: humans and LLMs fluidly communicate in natural language, and agents can display human-like reasoning and planning.
They can also be configured to emulate distinct cognitive and social behaviors, particularly through role playing, elevating their status from tools to collaborators.
We offer the following questions that explore dimensions of scientific teamwork which impact utility:
\begin{itemize}
    \item Trust is essential in human-AI interaction yet is dynamic and multifaceted. How should we design interactions to establish and maintain trust between humans and agent teammates in scientific contexts, and how does trust impact utility $\texttt{U}(H,A)$? 
    \item What is the relationship between AI over-reliance and utility $\texttt{U}(H,A)$ (Section~\ref{sec:risks:existential}), and how should we design for interactions that minimize over-reliance risks?
    \item Does integrating agents into real-time collaborative processes via asynchronous communication (e.g., AI processing delays~\citep{liu2024ai}) and mixed-initiative interaction~\citep{horvitz1999principles} increase \texttt{CA} at a reasonable increase in \texttt{CD}?
\end{itemize}




\section{Alternative Views}\label{sec:alternatives}

The primary opposing view to our core position is that a HAS perspective may provide little practical value for scientific teams once full costs are tallied.
Even if utility improves for some settings, the evaluation complexity and cost (not captured by collaboration disadvantage $\texttt{CD}$) may offset benefits.
\emph{Centaur evaluations}~\citep{haupt2025position} that compare HAS against humans and agents alone are important yet currently expensive and difficult to reproduce.
The idea is that the HAS is justified only when $\texttt{U}(H,A) -\max(\texttt{U}(H,H), \texttt{U}(A,A))$ is clearly large enough to exceed measurement costs.

We acknowledge the real challenge of cost-effective measurement of HAS in scientific discovery, particularly given the open-ended and long-horizon nature of these tasks.
However, a reasonable path is staged evaluation: start with lower cost evaluations (surveys and questionnaires, focused unit-tests of individual scientific collaborative capabilities~\citep{somasekharan2026sciconvbench}, observational data like transcripts of agent logs and aggregate statistics, and case studies), then invest in centaur studies when signals are strong.
We also propose the adoption of lightweight proxy metrics aligned with our utility model. 
To measure $\texttt{CD}$, researchers can evaluate \textbf{Review Time} as a cognitive load proxy, where a judge LLM measures how concise and actionable are the agent's questions and how long and meandering are research artifacts, as well as \textbf{Intervention Density}, which measures edit distance between agent artifacts and a human's finalized version. To measure $\texttt{CA}$, we advocate for \textbf{Simulated Oracle Synergy}, inspired by prior work on simulated feedback in multi-turn settings~\citep{wang2024mint}.
Here a stronger LLM acts as a proxy human scientist, allowing researchers to automatically measure how effectively an agent ingests and applies critical feedback.

\section{Conclusion}\label{sec:conclusion}
As we deploy AI agents into scientific teams, we should study these agents not just as solving an isolated optimization problem but also as a member of a \emph{human–agent system}. 
This lens centers research on sustaining human participation throughout discovery workflows, enabling fine- and coarse-grained intervention beyond artifact review and fixed gates, and offers concrete pathways to mitigate near-term risks while cultivating human–AI synergy in science. 
Our evidence motivates a mathematical framework which encourages exploring interaction modes that maximize the collaboration advantage or lower coordination costs. Studying AI Scientists as HAS is a pragmatic route to achieving safe and effective human–agent co-discovery.


\section*{Acknowledgments}

This work was authored in part by the National Laboratory of the Rockies for the U.S. Department of Energy (DOE), operated under Contract No. DE-AC36-08GO28308. It was also supported in part by the Pacific Northwest National Laboratory, which is operated by Battelle Memorial Institute for the U.S. Department of Energy under Contract DE-AC05–76RLO1830. This material is based upon work supported by the U.S. Department of Energy, Office of Science, ASCR under Award Number DE-SC0025425. Any opinions, findings, and conclusions or recommendations expressed in this material are those of the author(s) and do not necessarily reflect the views of the DOE, the United States Government or any agency thereof. The U.S. Government retains and the publisher, by accepting the article for publication, acknowledges that the U.S. Government retains a nonexclusive, paid-up, irrevocable, worldwide license to publish or reproduce the published form of this work, or allow others to do so, for U.S. Government purposes. This paper has been cleared by PNNL for public release as PNNL-SA-219899.

\bibliography{main}
\bibliographystyle{colm2026_conference}



\end{document}